\documentclass[11pt]{article}

\usepackage[preprint]{acl}
\usepackage{multirow}
\usepackage{booktabs}
\usepackage{placeins}
\usepackage{caption}
\usepackage{amsmath,amssymb,mathtools}
\usepackage{algorithm}
\usepackage{algpseudocode}
\usepackage{times}
\usepackage[table]{xcolor}
\usepackage{latexsym}
\usepackage{graphicx}
\usepackage{subcaption}
\usepackage{enumitem}
\usepackage[T1]{fontenc}
\usepackage{amsmath}
\usepackage[utf8]{inputenc}
\usepackage{arydshln}
\usepackage{hhline}
\usepackage{microtype}
\usepackage{listings}
\usepackage{inconsolata}

\usepackage[most]{tcolorbox}
\usepackage{xcolor}
\usepackage{amsmath}
\usepackage{amssymb}

\definecolor{casebg}{HTML}{F3F4F6}
\definecolor{caseframe}{HTML}{9CA3AF}
\definecolor{casebar}{HTML}{D1D5DB}
\definecolor{casetitle}{HTML}{111827}

\definecolor{errfg}{HTML}{B91C1C}
\definecolor{fixfg}{HTML}{B45309}
\definecolor{goodfg}{HTML}{15803D}
\definecolor{agreementfg}{HTML}{1D4ED8}

\newcommand{\badspan}[1]{\textcolor{errfg}{\textbf{#1}}}
\newcommand{\fixspan}[1]{\textcolor{fixfg}{\textbf{#1}}}

\newcommand{\badm}[1]{\textcolor{errfg}{#1}}
\newcommand{\fixm}[1]{\textcolor{fixfg}{#1}}
\newcommand{\goodm}[1]{\textcolor{goodfg}{#1}}

\newtcolorbox{caseblock}[2]{
  enhanced,
  breakable,
  colback=casebg,
  colframe=caseframe,
  colbacktitle=casebar,
  coltitle=casetitle,
  boxrule=0.5pt,
  arc=1.5mm,
  left=1.8mm,
  right=1.8mm,
  top=1.1mm,
  bottom=1.1mm,
  fontupper=\scriptsize\raggedright,
  title={\textbf{#1} \hfill #2},
  fonttitle=\small\bfseries
}

\newtcolorbox{compactproblem}{
  enhanced,
  colback=casebg,
  colframe=caseframe,
  boxrule=0.3pt,
  arc=1mm,
  boxsep=0.45mm,
  left=0.9mm,
  right=0.9mm,
  top=0.45mm,
  bottom=0.45mm,
  before skip=0pt,
  after skip=0pt,
  fontupper=\footnotesize\raggedright
}

\newtcolorbox{rolloutcard}[4]{
  enhanced,
  colback=#4!3,
  colframe=#4!60!black,
  boxrule=0.35pt,
  leftrule=0.9pt,
  arc=0.9mm,
  boxsep=0.45mm,
  left=0.9mm,
  right=0.9mm,
  top=0.45mm,
  bottom=0.45mm,
  toptitle=0.45mm,
  bottomtitle=0.45mm,
  before skip=0pt,
  after skip=0pt,
  fontupper=\footnotesize\raggedright,
  title={\textbf{#1}\hfill #2},
  before upper={%
    \noindent\textcolor{agreementfg}{\textbf{Rollout agreement}}%
    \hfill{\large\bfseries\textcolor{agreementfg}{#3}}%
    \par\vspace{0.3mm}%
    \noindent\rlap{\textcolor{casebar}{\rule{\linewidth}{1.2mm}}}%
    \textcolor{agreementfg}{\rule{#3\linewidth}{1.2mm}}%
    \par\vspace{0.6mm}%
  },
  colbacktitle=#4!8,
  coltitle=black,
  fonttitle=\footnotesize\bfseries,
  boxed title style={
    colback=#4!8,
    colframe=#4!60!black,
    boxrule=0pt,
    sharp corners
  }
}

\newtcolorbox{reasonbox}[1]{
  enhanced,
  colback=graybox,
  colframe=grayframe,
  boxrule=0.5pt,
  arc=1.5mm,
  left=1.2mm,
  right=1.2mm,
  top=1mm,
  bottom=1mm,
  fonttitle=\bfseries\small,
  fontupper=\footnotesize,
  title=#1
}

\lstdefinestyle{promptstyle}{
  basicstyle=\ttfamily\small,
  breaklines=true,
  columns=fullflexible,
  keepspaces=true,
  showstringspaces=false
}

\newtcblisting{promptbox}[2][]{
  colback=gray!5,
  colframe=gray!50,
  listing only,
  listing options={style=promptstyle},
  title={#2},
  sharp corners,
  boxrule=0.5pt,
  left=1mm,
  right=1mm,
  top=1mm,
  bottom=1mm,
  #1
}

\newcommand{\Ours}{Rollout-Agreement Reward}
\newcommand{\ours}{\textsc{RoAR}}

\newtcbox{\hlprimarytab}{on line, rounded corners, box align=base, colback=red!10,colframe=white,size=fbox,arc=3pt, before upper=\strut, top=-2pt, bottom=-4pt, left=-2pt, right=-2pt, boxrule=0pt}
\newtcbox{\hlsecondarytab}{on line, box align=base, colback=blue!10,colframe=white,size=fbox,arc=3pt, before upper=\strut, top=-2pt, bottom=-4pt, left=-2pt, right=-2pt, boxrule=0pt}
\newtcbox{\hlthirdtab}{on line, box align=base, colback=gray!10,colframe=white,size=fbox,arc=3pt, before upper=\strut, top=-2pt, bottom=-4pt, left=-2pt, right=-2pt, boxrule=0pt}
\newtcbox{\hlfourthtab}{on line, box align=base, colback=blue!20,colframe=white,size=fbox,arc=3pt, before upper=\strut, top=-2pt, bottom=-4pt, left=-2pt, right=-2pt, boxrule=0pt}

\title{Improving Reasoning with Rollout Agreement for Group-Based Policy Optimization}

\author{Sohyun Son \and Yerim Hong \and Jaeyun Park \and
  Mujeen Sung\thanks{Corresponding author: \texttt{mujeensung@khu.ac.kr}.} \\
  Kyung Hee University, Yongin, Republic of Korea \\
  \texttt{\{sohy, yrhong, jaeyunpark, mujeensung\}@khu.ac.kr}}

\begin{document}
\maketitle

\begin{abstract}
Recent methods for Reinforcement Learning with Verifiable Rewards (RLVR) use rollout diversity to improve reasoning performance or training efficiency.
Despite these gains, the extent to which differences among correct-answer rollouts reflect useful alternative reasoning or unresolved errors remains poorly understood.
We investigate this question through a blind paired analysis of a model trained with Group Relative Policy Optimization (GRPO), comparing the highest- and lowest-agreement correct-answer traces for the same problem.
Across 126 pairs, low-agreement traces have more spurious reasoning than high-agreement traces (35\% vs.\ 22\%), including internal inconsistencies and misuse of problem conditions.
Based on these findings, we propose \ours{} (\Ours{}), which adds reasoning-level agreement to GRPO's rewards for verified-correct rollouts.
\ours{} computes this signal from existing on-policy rollout groups, requiring no human-written rationales, additional sampling, or separately trained process evaluator.
On five mathematical reasoning benchmarks, the 1.5B GRPO-\ours{} model reaches 55.5\% mean pass@1 under zero-shot evaluation using 7K training examples, exceeding GRPO and GRPO-CARE by 6.0 and 5.5 percentage points.
A separate blind audit of 100 problems answered correctly by all four 1.5B models finds spurious rates of 18\% for GRPO-\ours{}, 32\% for GRPO, 27\% for GRPO-CARE, and 37\% for DRA-GRPO.
Our findings identify reasoning validity as an important consideration when using rollout differences to guide learning.
Our code is available at \url{https://github.com/ldi-kyunghee/RoAR}.
\end{abstract}
\begin{figure}[!t]
  \centering
  \includegraphics[width=\linewidth]{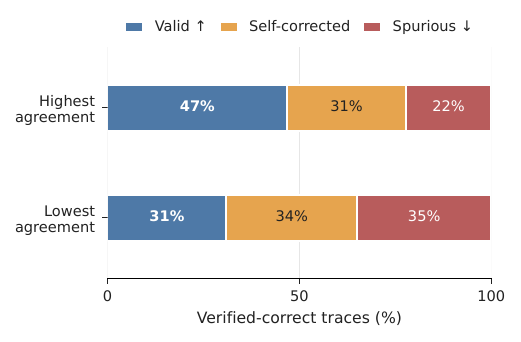}
  \caption{
  \textbf{Different traces can reach the same correct answer.}
  Blind paired analysis of the highest- and lowest-agreement verified-correct traces for each of 126 problems from a GRPO-trained 1.5B model.
  Low-agreement traces have a higher spurious rate (35\% vs.\ 22\%) and a lower valid rate (31\% vs.\ 47\%).
  }
  \label{fig:motivation}
\end{figure}

\section{Introduction}

Reinforcement Learning with Verifiable Rewards (RLVR) trains language models to generate intermediate reasoning using automatically checked outcomes, avoiding manual evaluation of each generated solution~\citep{guo2025deepseekr1, lambert2025tulu3, shao2024deepseekmath}.
Group Relative Policy Optimization (GRPO) samples multiple candidate solutions, or rollouts, per problem and updates the policy using their relative rewards~\citep{shao2024deepseekmath}.
Recent methods use differences among these rollouts to improve reasoning performance or training efficiency.
DRA-GRPO~\citep{chen2025dragrpo} reweights rewards to emphasize less redundant responses, while MMR-GRPO~\citep{wei2026mmrgrpo} combines outcome rewards and diversity when reweighting rollout rewards.

The value of this preference depends on what makes a rollout different from its peers.
A distinct reasoning path may offer a useful alternative, but it may also depart from others through faulty operations, misread conditions, or unresolved contradictions.
Final-answer verification leaves this distinction open: a correct answer can follow from unsound intermediate reasoning~\citep{zhang2025lessonsdevelopingprocessreward, chen2025grpocare, wang2025scs}.
This raises a question: \emph{what do differences among verified-correct rollouts capture, and how should they guide learning?}

We investigate this question by comparing reasoning traces that reach the same verified answer for the same problem.
In a blind paired analysis of a GRPO-trained 1.5B model, the lowest-agreement traces have a higher spurious-reasoning rate than the highest-agreement traces (35\% vs.\ 22\%; Figure~\ref{fig:motivation}).
The differences include internal inconsistencies, faulty operations, and misuse of problem conditions (Section~\ref{sec:motivation}).
These results associate agreement with differences in reasoning validity that final-answer verification leaves unresolved.

We propose \ours{} (\Ours{}), which adds a reasoning-level agreement reward to GRPO.
It measures each trace's agreement with the full on-policy group and awards the bonus only to verified-correct rollouts.
\ours{} computes this signal from existing rollouts, requiring no human-written rationales, additional sampling, or separately trained process evaluator.

We evaluate the resulting policy through both answer accuracy and reasoning validity.
GRPO-\ours{} achieves 55.5\% mean pass@1 across five mathematical reasoning benchmarks at 1.5B, exceeding GRPO and GRPO-CARE by 6.0 and 5.5 percentage points, respectively (Table~\ref{tab:main_table}).
A separate blind audit of 100 problems answered correctly by all four 1.5B models finds fewer spurious traces under GRPO-\ours{} than under GRPO, GRPO-CARE, and DRA-GRPO (Section~\ref{sec:qual_analysis}).

Our contributions are threefold:
\begin{itemize}[leftmargin=*, itemsep=0.25em, topsep=0.25em, parsep=0pt]
    \item We analyze reasoning validity among verified-correct rollouts and identify an association between low agreement and unresolved errors.
    \item We introduce \ours{}, an agreement reward computed from existing rollout groups, and evaluate its accuracy at two model scales.
    \item We compare reasoning validity across GRPO, GRPO-\ours{}, GRPO-CARE, and DRA-GRPO on shared correct-answer problems, alongside an analysis of final-answer consensus.
\end{itemize}

\section{What Do Differences Among Correct-Answer Rollouts Capture?}

\label{sec:motivation}

We examine what differences among rollouts reflect when final-answer verification assigns them the same correctness reward.
Comparing verified-correct traces for the same problem lets us investigate variation in reasoning while holding the problem and final-answer correctness fixed.
We ask whether traces that differ more from their peers are more likely to contain unresolved errors.

We sample $K=32$ rollouts per problem on MinervaBench~\citep{lewkowycz2022solving} from a GRPO-trained DeepSeek-R1-Distill-Qwen-1.5B model~\citep{guo2025deepseekr1} and compute agreement among verified-correct rollouts.
Following prior work on reasoning-level consistency~\citep{knappe2025semantic, wan2025rasc, hong2025slim, ling2026correct}, we combine semantic similarity and lexical overlap.
For each verified-correct rollout, we average its pairwise similarities to the other verified-correct rollouts for the same problem to obtain its agreement score.
For each problem with at least five verified-correct rollouts, we pair the highest- and lowest-agreement traces, yielding 126 anonymized pairs.
The traces are annotated with agreement labels hidden, using the guideline in Appendix~\ref{app:qual_prompt}.
A \textit{valid} trace gives sound support for the answer.
A \textit{self-corrected} trace explicitly repairs or abandons an intermediate error before reaching a sound final path.
A \textit{spurious} trace contains unresolved errors or insufficient support for the answer.
We also annotate reasoning-error types.
Generation settings are provided in Appendix~\ref{app:motivation_generation}.

The lowest-agreement traces have a higher spurious rate than the highest-agreement traces (35\% vs.\ 22\%) and a lower valid rate (31\% vs.\ 47\%; Figure~\ref{fig:motivation}).
Even after fixing final-answer correctness, divergence from peer traces is therefore associated with differences in reasoning validity.
Both agreement groups include valid and spurious traces.

\begin{figure}[t]
  \centering
  \includegraphics[width=\linewidth]{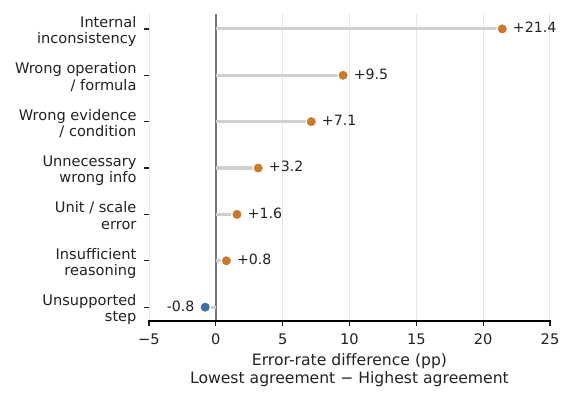}
  \caption{\textbf{Internal inconsistency shows the largest error-rate gap between agreement groups.}
  Error-rate differences in the 126 paired comparisons are computed as lowest agreement minus highest agreement.
  Positive values indicate higher error rates in the lowest-agreement verified-correct traces.}
  \label{fig:motivation-errors}
\end{figure}

The largest error-rate gaps occur in internal inconsistency (+21.4 percentage points), wrong operation or formula (+9.5), and wrong evidence or condition (+7.1; Figure~\ref{fig:motivation-errors}).
These failures include conflicting reasoning steps, faulty computations, and misuse of the given conditions.
They provide concrete sources of variation among traces that final-answer verification treats equally.

These paired comparisons link rollout agreement to reasoning-validity differences among outputs receiving the same correctness reward.

\section{Rollout-Agreement Reward}

\label{sec:roar}

\begin{figure*}[t]
  \centering
  \includegraphics[width=\textwidth]{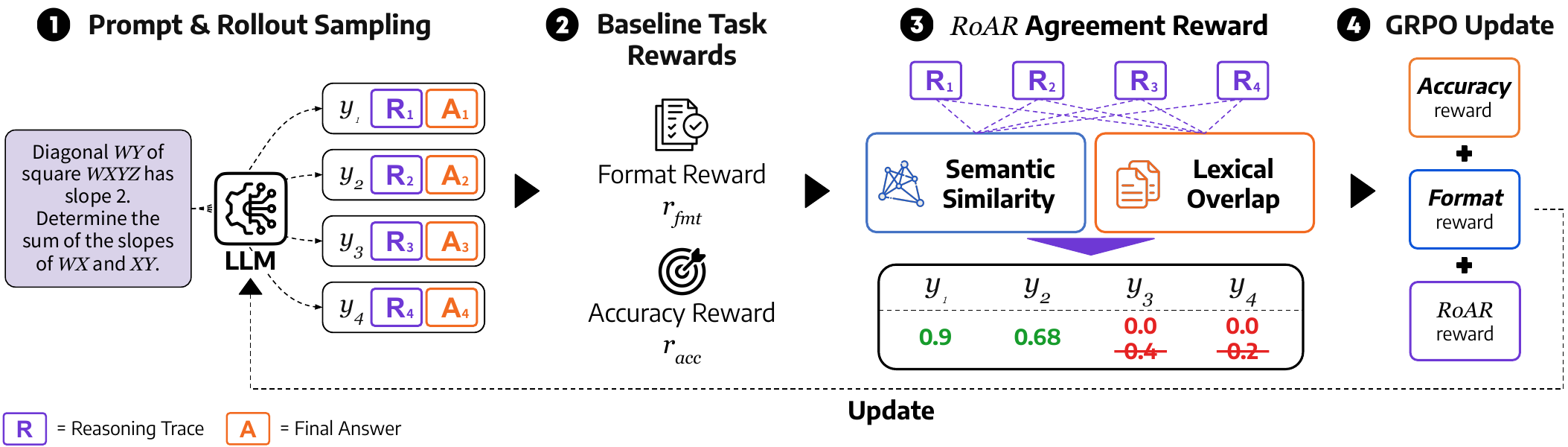}
  \caption{Overview of \ours{} within group-based policy optimization.
  \ours{} measures reasoning-level agreement over the sampled rollout group and assigns the agreement reward only to verified-correct rollouts before the group-relative policy update.
  \textbf{R} and \textbf{A} denote the \textit{reasoning trace} and \textit{final answer}; the figure uses $K=4$.}
  \label{fig:framework}
\end{figure*}

\ours{} adds a reasoning-level agreement reward to GRPO.
It uses the full on-policy rollout group as references and awards the bonus only to verified-correct rollouts.
This favors correct-answer traces that share reasoning patterns with other rollouts sampled for the same prompt, without additional rollout sampling.
Diversity-based reward adjustment emphasizes less redundant responses~\citep{chen2025dragrpo, wei2026mmrgrpo}; \ours{} uses agreement with peer traces as an additional learning signal.

\subsection{Measuring Rollout Agreement}
\label{subsec:roar_reward}

We quantify rollout agreement over the reasoning traces extracted from the \texttt{<think>} spans.
For a prompt $x$, let $\{y_i\}_{i=1}^K$ denote the group of rollouts sampled from the policy, and let $t_i$ denote the reasoning trace of rollout $y_i$.
To obtain an agreement metric, we combine semantic similarity and lexical overlap, following prior work that treats these two views as complementary signals for reasoning-level agreement~\citep{knappe2025semantic, wan2025rasc, hong2025slim, ling2026correct}.
Specifically, we define the pairwise similarity score $s(i,j)$ between two reasoning traces as
\begin{equation}
s(i,j)
=
\alpha\, s_{\text{sem}}(i,j)
+
(1-\alpha)\, s_{\text{lex}}(i,j),
\label{eq:s_mix}
\end{equation}
where $s_{\text{sem}}(i,j)$ is the cosine similarity between embeddings of the reasoning traces $t_i$ and $t_j$, and $s_{\text{lex}}(i,j)$ is token-level F1 between the two traces.
The coefficient $\alpha \in [0,1]$ controls the trade-off between semantic similarity and lexical overlap.

For each rollout, we compute its agreement score as the average pairwise similarity with all other rollouts sampled for the same prompt.
\begin{equation}
a_i^{\text{all}}
=
\frac{1}{K-1}
\sum_{\substack{j=1 \\ j\ne i}}^K
s(i,j).
\label{eq:all_ref_agreement}
\end{equation}
The diagnostic in Section~\ref{sec:motivation} compares correct-answer traces to control for answer correctness.
For training, we use the full on-policy group as references, keeping the score well-defined even when verified-correct rollouts are sparse or absent.
This group represents the reasoning patterns sampled by the current policy for the prompt.
We evaluate the reference-set choice in Section~\ref{sec:ablation}.

Similarity can also be high among rollouts with incorrect final answers.
We therefore award the agreement bonus only to verified-correct rollouts.
Let $\mathbb{I}_{\text{correct}}(y_i)$ be an indicator for whether the final answer of rollout $y_i$ is correct.
The \ours{} reward for rollout $y_i$ is defined as
\begin{equation}
r_{\text{\ours}}(y_i)
=
\mathbb{I}_{\text{correct}}(y_i)
\cdot
a_i^{\text{all}}.
\label{eq:roar_reward}
\end{equation}
Thus, an incorrect rollout receives no agreement reward, even if its trace is similar to other rollouts.

\subsection{Integration into Group-based Policy Optimization}
\label{subsec:roar_grpo}

\ours{} is integrated into GRPO by adding an agreement term to the per-rollout reward.
For each rollout, we first compute the standard task reward used in our GRPO baseline, consisting of a format reward for output-structure compliance and a binary accuracy reward for verifiable final-answer correctness.
We then add the rollout-agreement reward defined above.
\begin{equation}
r_i
=
w_{\text{fmt}} r_{\text{fmt}}(y_i)
+
w_{\text{acc}} r_{\text{acc}}(y_i)
+
w_{\text{\ours}} r_{\text{\ours}}(y_i).
\label{eq:total_reward}
\end{equation}
where $w_{\text{fmt}}$, $w_{\text{acc}}$, and $w_{\text{\ours}}$ are the weights for the format, accuracy, and rollout-agreement rewards, respectively.

Given the rollout rewards $\{r_i\}_{i=1}^K$, we compute group-relative advantages within each prompt group and apply the standard GRPO update~\citep{shao2024deepseekmath}.
Thus, \ours{} preserves the existing GRPO optimization procedure and can be incorporated as a plug-in reward.
It provides reasoning-level feedback through the existing rollout rewards without a separately trained reasoning-quality evaluator.
Reward weights and training hyperparameters are provided in Appendix~\ref{app:hyperparams}.

\paragraph{Ablation variants.}
By default, \ours{} computes agreement relative to all rollouts in the on-policy group and assigns the resulting reward only to verified-correct rollouts.
To isolate these choices, we test a reference-set ablation that uses only verified-correct rollouts as references, and a reward-target ablation that assigns the agreement reward to all rollouts regardless of correctness.

\section{Experiments}

We evaluate answer accuracy, reward construction, and results across model scales and training seeds.
Section~\ref{sec:analysis} examines reasoning validity and final-answer consensus.

\begin{table*}[t]
\centering
\small
\setlength{\tabcolsep}{4.5pt}
\renewcommand{\arraystretch}{1.05}
\begin{tabular}{lcccccc}
\toprule
\textbf{Model} & \textbf{AIME26} & \textbf{MATH-500} & \textbf{AMC23} & \textbf{Minerva} & \textbf{OlympiadBench} & \textbf{Avg.} \\
\midrule
DeepSeek-R1-Distill-Qwen-1.5B & 20.0 & 82.8 & 62.9 & 26.5 & 43.3 & 47.1 \\
OpenRS3 \citep{dang2026openrs}& \textbf{26.7} & 84.4 & 72.5 & 26.8 & 51.3 & 52.3 \\
DRA-GRPO \citep{chen2025dragrpo}& 23.3 & 86.2 & 75.0 & \textbf{32.4} & 53.0 & 54.0 \\
\midrule
GRPO$^\dagger$ & 16.7 & 84.4 & 65.0 & 27.9 & 53.5 & 49.5 \\
GRPO-CARE$^\dagger$ \citep{chen2025grpocare} & 20.0 & 82.0 & 65.0 & 29.0 & 54.2 & 50.1 \\
GRPO w/ LLM reward$^\dagger$ \citep{corbitt2025ruler} & 20.0 & 83.2 & 75.0 & 29.4 & 51.6 & 51.8 \\
\rowcolor{gray!15}
GRPO-\ours{}$^\dagger$ (Ours)
& \textbf{26.7} & 84.8 & \textbf{80.0} & 30.9 & \textbf{55.3} & \textbf{55.5} \\

\quad Reference set: All $\rightarrow$ Correct
& 20.0 & \textbf{86.4} & 70.0 & 31.6 & 53.0 & 52.2 \\

\quad Reward target: Correct $\rightarrow$ All
& 23.3 & 82.2 & 75.0 & 27.9 & 51.1 & 51.9 \\

\bottomrule
\end{tabular}
\caption{
Pass@1 comparison across mathematical reasoning benchmarks.
Numbers are in \%.
Bold numbers indicate the highest value in each column.
$\dagger$ denotes models trained by us.
All reported AIME26 scores are evaluated by us; external-baseline scores on the other four benchmarks are taken from~\citet{dang2026openrs} and~\citet{chen2025dragrpo}. Indented rows are ablations of \ours{}.
}
\label{tab:main_table}
\end{table*}

\subsection{Experimental Setup}
\label{sec:experiments}

\paragraph{Training data.}
We use the 7K mathematical reasoning training set from Open-RS~\citep{dang2026openrs}, which consists of problems selected from the s1 and DeepScaleR data sources~\citep{muennighoff2025s1, luo2025deepscaler}.
It spans a range of difficulties and has been used in low-resource GRPO training for small reasoning models~\citep{dang2026openrs, chen2025dragrpo}, allowing us to isolate the effect of the rollout-agreement reward under a fixed data budget.

\paragraph{Evaluation benchmarks.}
We evaluate on five mathematical reasoning benchmarks:
AIME26\footnote{\url{https://huggingface.co/datasets/math-ai/aime26}},
MATH-500~\citep{lightman2023lets, hendrycks2021measuring},
AMC23\footnote{\url{https://huggingface.co/datasets/knoveleng/AMC-23}},
Minerva~\citep{lewkowycz2022solving},
and OlympiadBench~\citep{he2024olympiadbench}.
Compared with the Open-RS evaluation suite, we replace AIME24 with AIME26 and evaluate all models with reported AIME26 scores under the same inference and answer-verification settings; the remaining benchmarks are unchanged.

\paragraph{Model and training.}
We use DeepSeek-R1-Distill-Qwen-1.5B~\citep{guo2025deepseekr1} as the base model.
Our implementation builds on the Open-RS training pipeline~\citep{dang2026openrs}.
We train an outcome-only GRPO baseline with format and binary accuracy rewards and reimplement GRPO-CARE~\citep{chen2025grpocare} under the same setup; \ours{} adds the rollout-agreement reward defined in Section~\ref{sec:roar}.
GRPO-CARE provides a comparison using reasoning--answer coherence.
We also train GRPO w/ LLM reward, which uses the RULER-style approach~\citep{corbitt2025ruler} to obtain trajectory feedback from an LLM judge within each rollout group.
For the 4B experiments, we train Qwen3-4B~\citep{qwen3technicalreport} with thinking mode enabled using the same 7K examples and training budget.

In the main experiments, we set the rollout-agreement weight to $w_{\text{\ours}}=1.0$ and the semantic--lexical mixing coefficient to $\alpha=0.5$.
We study the sensitivity to these choices in Section~\ref{sec:agreement_sensitivity} and Appendices~\ref{app:alpha_ablation} and~\ref{app:weight_ablation}.
Implementation details for the GRPO-CARE and GRPO w/ LLM reward baselines are provided in Appendix~\ref{app:hyperparams}.

\paragraph{Evaluation protocol.}
We report pass@1 under zero-shot prompting.
For pass@1, the model generates one solution per problem, and the final answer is extracted and checked against the ground-truth answer.
For pass@4, we sample four solutions per problem and count the problem as correct if any extracted final answer is verified correct.
For models trained in our environment, including GRPO, GRPO-CARE, GRPO w/ LLM reward, \ours{}, and our ablation variants, we use our reimplementation of the Open-RS-style answer-extraction and verification protocol.

\subsection{Main Results}
\label{sec:main_results}

Table~\ref{tab:main_table} reports pass@1 across five mathematical reasoning benchmarks.
Performance differences are computed from unrounded scores.
Our main GRPO-\ours{} model achieves the highest average pass@1 of 55.5\% among the 1.5B models.
Compared with the base DeepSeek-R1-Distill-Qwen-1.5B model, our GRPO-\ours{} method improves average pass@1 by 8.4 points.
GRPO-\ours{} outperforms GRPO by 6.0 points, GRPO-CARE by 5.5 points, and GRPO w/ LLM reward by 3.7 points.
Paired bootstrap intervals over evaluation problems show positive pass@1 gains over GRPO and GRPO-CARE, with 95\% CIs of [1.6, 10.6] and [0.9, 10.0] points, respectively (Appendix~\ref{app:bootstrap_ci}).

Table~\ref{tab:main_table} also compares prior 1.5B reasoning models, with externally reported scores identified in the caption.
GRPO-\ours{} exceeds the listed OpenRS3 and DRA-GRPO averages by 3.2 and 1.5 points, respectively.
GRPO-\ours{} achieves the best scores on AMC23 and OlympiadBench, at 80.0\% and 55.3\%.
OpenRS3 matches its score on AIME26, while DRA-GRPO scores slightly higher on MATH-500 and Minerva.


\begin{table}[t]
  \centering
  \small
  \setlength{\tabcolsep}{8pt}
  \begin{tabular}{lc}
    \toprule
    \textbf{Method} & \textbf{Pass@4} \\
    \midrule
    GRPO
      & 66.2 \\
    GRPO-CARE
      & 65.1 {\tiny -1.1} \\
    DRA-GRPO
      & 63.7 {\tiny -2.5} \\
    GRPO-\ours{} (\textbf{Ours})
      & \textbf{68.3} {\tiny \textbf{+2.1}} \\
    \bottomrule
  \end{tabular}
  \caption{Macro-average pass@4 across five benchmarks. Numbers are in \%. The small values denote percentage-point differences from GRPO. DRA-GRPO results are obtained from our evaluation of its released checkpoint.}
  \label{tab:pass4}
\end{table}

\begin{table*}[t]
\centering
\small
\setlength{\tabcolsep}{4pt}
\renewcommand{\arraystretch}{1.05}
\begin{tabular}{lrrrrrr}
\toprule
& \multicolumn{3}{c}{\textbf{1.5B}} & \multicolumn{3}{c}{\textbf{4B}} \\
\cmidrule(lr){2-4}\cmidrule(lr){5-7}
\textbf{Task} & \textbf{GRPO} & \textbf{GRPO-\ours{}} & \textbf{$\Delta$} & \textbf{GRPO} & \textbf{GRPO-\ours{}} & \textbf{$\Delta$} \\
\midrule
AIME26 & $22.2{\pm}6.9$ & $\mathbf{24.4{\pm}3.8}$ & +2.2 & $61.1{\pm}5.1$ & $\mathbf{66.7{\pm}5.8}$ & +5.6 \\
MATH-500 & $83.8{\pm}0.7$ & $\mathbf{83.9{\pm}1.0}$ & +0.1 & $95.7{\pm}0.4$ & $\mathbf{96.4{\pm}0.4}$ & +0.7 \\
AMC23 & $70.8{\pm}8.0$ & $\mathbf{74.2{\pm}5.2}$ & +3.3 & $\mathbf{96.7{\pm}2.9}$ & $95.8{\pm}5.2$ & $-0.8$ \\
Minerva & $29.4{\pm}1.3$ & $\mathbf{30.5{\pm}2.0}$ & +1.1 & $46.2{\pm}0.8$ & $\mathbf{46.4{\pm}0.4}$ & +0.2 \\
OlympiadBench & $51.9{\pm}1.4$ & $\mathbf{53.8{\pm}1.3}$ & +1.9 & $73.8{\pm}0.6$ & $\mathbf{75.4{\pm}0.6}$ & +1.5 \\
Average & $51.6{\pm}1.9$ & $\mathbf{53.4{\pm}2.1}$ & \textbf{+1.7} & $74.7{\pm}1.3$ & $\mathbf{76.1{\pm}1.9}$ & \textbf{+1.4} \\
\bottomrule
\end{tabular}
\caption{Pass@1 (\%), reported as mean $\pm$ sample SD over three training seeds for DeepSeek-R1-Distill-Qwen-1.5B and Qwen3-4B. Within each model size, both methods use the same 7K examples, training budget, and evaluation protocol. Average is the five-benchmark macro-average; its SD is computed across the per-seed macro-averages. $\Delta$ is the GRPO-\ours{} improvement in percentage points, computed from unrounded scores.}
\label{tab:rebuttal_performance}
\end{table*}

As shown in Table~\ref{tab:pass4}, GRPO-\ours{} achieves 68.3\% pass@4, 2.1 points higher than outcome-only GRPO.
GRPO-CARE obtains 65.1\%, which is 1.1 points below the GRPO baseline in this setting.
Our evaluation of the released DRA-GRPO checkpoint yields 63.7\% mean pass@4, 4.7 percentage points below GRPO-\ours{}.
Paired bootstrap confidence intervals for the pass@4 comparisons with GRPO, GRPO-CARE, and DRA-GRPO are reported in Appendix~\ref{app:bootstrap_ci}.

GRPO-\ours{} takes nearly the same wall-clock time as outcome-only GRPO (Appendix~\ref{app:resource_overhead}).
The GRPO w/ LLM reward run takes 21 hours 58 minutes and incurs \$42.97 in judge API costs.
The GRPO-\ours{} run takes 11 hours 16 minutes without external judge calls.

\subsection{Ablation Study}
\label{sec:ablation}

We examine how the reference set used to compute agreement and the choice of rollouts that receive the reward affect performance (Table~\ref{tab:main_table}).
The default configuration uses all rollouts as references but rewards only verified-correct rollouts, achieving 55.5\% mean pass@1.
Rewarding all rollouts lowers this score to 51.9\%, while restricting the reference set to verified-correct rollouts lowers it to 52.2\%.
In this setting, the full rollout group provides a useful reference, while conditioning the reward on correctness improves performance.

\subsection{Performance Across Model Scales and Training Seeds}
\label{sec:qwen4b}

To assess training-run variability, we repeat the central comparison between GRPO and GRPO-\ours{} with training seeds 0, 7, and 2025 at both model sizes.
The models trained by us in Table~\ref{tab:main_table} use training seed 2025, including the reward-design ablations.
Table~\ref{tab:rebuttal_performance} reports the per-benchmark results at both model sizes.
At 1.5B, GRPO-\ours{} improves the mean pass@1 on all five benchmarks, raising the five-benchmark average from $51.6\pm1.9$ to $53.4\pm2.1$.

To test whether the same reward remains useful with a larger model, we train Qwen3-4B under the same data, training budget, and evaluation protocol.
Across three training seeds, GRPO-\ours{} improves the mean pass@1 on four of five benchmarks and raises the five-benchmark average from 74.7\% to 76.1\% (Table~\ref{tab:rebuttal_performance}).
The gains include 0.7 points on MATH-500 and 1.5 points on OlympiadBench, while AMC23 decreases by 0.8 points on average.

The five-benchmark averages for individual training seeds are reported in Appendix~\ref{app:seed_results}.

\section{Analysis}

\label{sec:analysis}

We examine whether the accuracy gains reported in Section~\ref{sec:main_results} are accompanied by changes in reasoning validity.
We then analyze final-answer consensus and sensitivity to the agreement score and reward weight.

\subsection{Reasoning Validity Beyond Correct Answers}
\label{sec:qual_analysis}

We compare the reasoning supporting correct answers from the 1.5B GRPO, GRPO-CARE, DRA-GRPO, and GRPO-\ours{} models.
The blind audit uses 100 shared problems answered correctly by all four models, holding the problem and final-answer correctness fixed across methods.
We shuffle the traces and annotate them with model identities hidden, using the three validity labels defined in Section~\ref{sec:motivation} and the guideline in Appendix~\ref{app:qual_prompt}.
For DRA-GRPO, we use rollouts generated in our environment from its released checkpoint\footnote{\url{https://huggingface.co/SpiceRL/DRA-GRPO}}.

\begin{figure}[t]
  \centering
  \includegraphics[width=\linewidth]{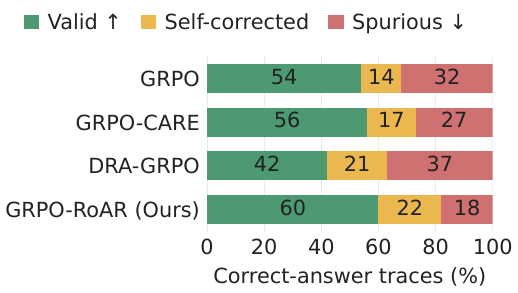}
  \caption{
  \textbf{Reasoning validity after training.}
  Blind human audit on 100 shared problems answered correctly by GRPO, GRPO-CARE, DRA-GRPO, and GRPO-\ours{} (1.5B).
  Bars show the percentages of valid, self-corrected, and spurious traces.
  }
  \label{fig:qual}
\end{figure}

Figure~\ref{fig:qual} shows spurious rates of 32\% for GRPO, 27\% for GRPO-CARE, 37\% for DRA-GRPO, and 18\% for GRPO-\ours{}.
GRPO-\ours{} also has the highest valid rate (60\%), while 22\% of its traces are self-corrected.
In this setting, the accuracy improvement under \ours{} is accompanied by fewer unresolved errors in correct-answer traces.
Human evaluation at the 4B scale also finds fewer spurious traces under GRPO-\ours{} than under GRPO; details are provided in Appendix~\ref{app:human_4b_qual}.
Appendix~\ref{app:case_study} illustrates the distinction between valid, self-corrected, and spurious traces.

As a complementary check, we apply GPT-6.1 Sol with low reasoning effort to the traces from the 1.5B audit using the same guideline.
Human--LLM inter-annotator agreement (IAA), measured as exact agreement across the three categories, is 79.25\%, with Cohen's $\kappa=0.665$.
The LLM judge also assigns GRPO-\ours{} the lowest spurious rate; the full counts and protocol are provided in Appendix~\ref{app:llm_qual}.

\subsection{Accuracy and Final-Answer Consensus}
\label{sec:consensus_conditioned}

Because \ours{} rewards reasoning-level agreement, a natural concern is that it encourages rollouts to converge to a single dominant final answer.
We examine whether increased final-answer consensus accounts for its accuracy gains.
For GRPO, GRPO-CARE, DRA-GRPO, and GRPO-\ours{}, we examine final-answer consensus among the four rollouts sampled per problem.
For DRA-GRPO, we evaluate the same released checkpoint used in the validity audit in our environment.
A problem has \textit{strong answer consensus} if at least three of the four rollouts produce the same extracted final answer, and \textit{weak answer consensus} otherwise.

\begin{table}[t]
\centering
\small
\setlength{\tabcolsep}{4.5pt}
\renewcommand{\arraystretch}{1.05}
\begin{tabular}{@{}lccc@{}}
\toprule
\textbf{Model} & \textbf{Strong} & \textbf{Weak Pass@1} & \textbf{Weak Pass@4} \\
\midrule
GRPO & 48.7 & 21.2 & 47.9 \\
GRPO-CARE & \textbf{52.5} & 14.8 & 43.0 \\
DRA-GRPO & 48.4 & 20.9 & 46.9 \\
GRPO-\ours{} & 48.8 & \textbf{28.1} & \textbf{52.4} \\
\bottomrule
\end{tabular}
\caption{
Answer-consensus diagnostic over four sampled rollouts.
Strong denotes the percentage of problems where at least three rollouts produce the same extracted final answer.
Weak Pass@1 and Weak Pass@4 are computed only on the weak-consensus subset.
Numbers are in \% and macro-averaged across five benchmarks.
DRA-GRPO results are obtained from our evaluation of its released checkpoint.
}
\label{tab:consensus_conditioned}
\end{table}

GRPO-\ours{} has nearly the same strong-consensus rate as GRPO (48.8\% vs. 48.7\%).
On each model's weak-consensus subset, GRPO-\ours{} reaches 28.1\% pass@1 and 52.4\% pass@4, compared with 21.2\% and 47.9\% for GRPO.
DRA-GRPO has a similar strong-consensus rate (48.4\%), with weak-consensus pass@1 and pass@4 of 20.9\% and 46.9\%, respectively.
These results suggest that \ours{}'s accuracy gains extend beyond increasing the frequency of strong final-answer consensus.
Its higher accuracy on weak-consensus subsets shows that improved performance also occurs where sampled answers remain dispersed.
Per-benchmark results are provided in Appendix~\ref{app:consensus_conditioned}.

\subsection{Sensitivity to Agreement Scoring and Reward Weight}
\label{sec:agreement_sensitivity}

We examine how the semantic--lexical mixing coefficient and rollout-agreement reward weight affect performance.

\begin{figure}[t]
  \centering
  \includegraphics[width=\linewidth]{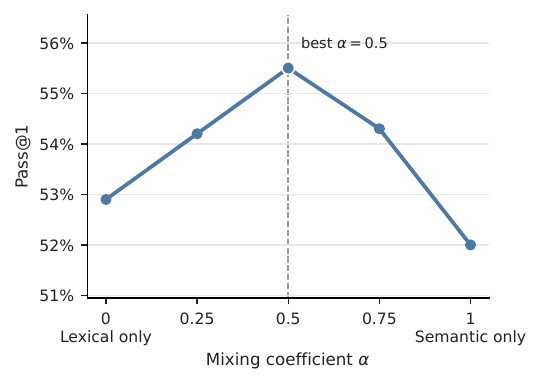}
  \caption{
  Ablation of the mixing coefficient $\alpha$ in the pairwise similarity score (Eq.~\ref{eq:s_mix}).
  We report mean pass@1 across the five benchmarks. Numbers are in \%.
  $\alpha=0$ denotes lexical-only agreement, and $\alpha=1$ denotes semantic-only agreement.
  }
  \label{fig:alpha}
\end{figure}

\paragraph{Semantic--lexical mixing.}
We ablate the mixing coefficient $\alpha$ in Eq.~\ref{eq:s_mix}, which interpolates between lexical overlap and semantic similarity.
Figure~\ref{fig:alpha} shows that the balanced setting $\alpha=0.5$ performs best among the tested settings, reaching 55.5\% mean pass@1, while lexical-only and semantic-only agreement obtain 52.8\% and 52.0\%.
This suggests that semantic similarity and lexical overlap provide complementary agreement signals.
We therefore use $\alpha=0.5$ in the main experiments.
Detailed per-benchmark results are provided in Appendix~\ref{app:alpha_ablation}.

\begin{figure}[t]
  \centering
  \includegraphics[width=\linewidth]{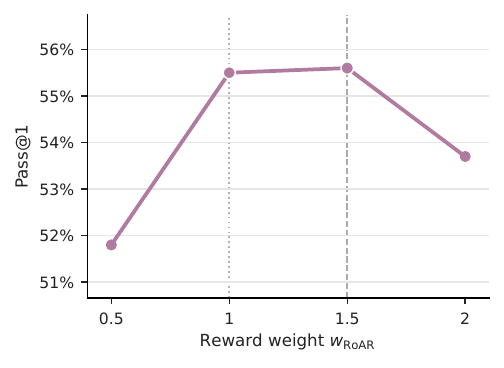}
  \caption{
  Sensitivity to the rollout-agreement reward weight $w_{\ours}$.
  We report mean pass@1 across the five benchmarks. Numbers are in \%.
  }
  \label{fig:weight_ablation}
\end{figure}

\paragraph{Rollout-agreement weight.}
Figure~\ref{fig:weight_ablation} shows similar performance at $w_{\ours}=1.0$ and $w_{\ours}=1.5$, with mean pass@1 scores of 55.5\% and 55.6\%, respectively.
Mean pass@1 falls to 51.8\% at $w_{\ours}=0.5$ and 53.7\% at $w_{\ours}=2.0$.
We use $w_{\ours}=1.0$ as the default.
Detailed per-benchmark results are provided in Appendix~\ref{app:weight_ablation}.

\section{Related Work}

\subsection{Reasoning-Aware Rewards in GRPO}

GRPO assigns the same correctness reward to correct-answer rollouts even when their reasoning quality differs~\citep{shao2024deepseekmath, lambert2025tulu3}.
Recent methods exploit properties of the rollout group to modify these rewards.
DRA-GRPO~\citep{chen2025dragrpo} reweights rewards according to semantic redundancy, giving greater weight to responses that are less similar to their peers.
MMR-GRPO~\citep{wei2026mmrgrpo} combines outcome rewards and semantic redundancy through Maximal Marginal Relevance to reweight rollout rewards and accelerate training.
These methods seek to give less redundant responses more influence during learning.
Our paired analysis examines the reasoning differences among verified-correct outputs, including errors that final-answer rewards leave unresolved.
\ours{} tests peer agreement as an additional signal in this setting.

Other approaches provide feedback on the reasoning process itself.
Process reward models score intermediate steps using an external evaluator~\citep{lightman2023lets, zhang2025lessonsdevelopingprocessreward, cheng2025stopsummationminformcredit}, while RULER uses an LLM judge to score trajectories within each rollout group~\citep{corbitt2025ruler}.

SCS~\citep{wang2025scs} and GRPO-CARE~\citep{chen2025grpocare} specifically address spurious reasoning with auxiliary consistency signals.
SCS uses trajectory agreement obtained through visual perturbation and truncation-resampling to down-weight unreliable traces in multimodal reasoning.
GRPO-CARE estimates reasoning--answer coherence with a reference model and incorporates it as a consistency reward.
\ours{} computes agreement directly within existing rollout groups and awards the bonus to verified-correct outputs, using neither input perturbations nor a separate reference model.

\subsection{Agreement Signals from Multiple Rollouts}

Self-consistency samples multiple reasoning paths and predicts an answer by majority vote~\citep{wang2023selfconsistency}.
Consistency across generations also supports hallucination detection~\citep{manakul2023selfcheckgpt} and confidence estimation~\citep{lyu2025calibrating}.
Reasoning-level methods compare paths or rationales for answer aggregation~\citep{knappe2025semantic, wan2025rasc}, while inter-chain similarity can reduce the sampling cost of self-consistency~\citep{hong2025slim}.

These approaches use agreement primarily at inference time.
\ours{} uses reasoning-level agreement during training.
It reuses the rollout groups generated by GRPO~\citep{shao2024deepseekmath} to compute an auxiliary reward and incorporate it into policy updates, without additional sampling.


\section{Conclusion}
\label{sec:conclusion}

We investigated what differences among verified-correct rollouts capture.
A blind paired analysis found more unresolved errors in low-agreement traces, including internal inconsistencies, faulty operations, and misuse of problem conditions.
This association motivates testing agreement as additional feedback among outputs receiving the same correctness reward.

\ours{} implements this design using existing on-policy rollout groups and rewards verified-correct traces that share reasoning patterns with their peers.
Across three training seeds, it improves average pass@1 at both 1.5B and 4B.
In the blind audit of 100 problems answered correctly by all four 1.5B models, its spurious rate is 18\%, compared with 32\% for GRPO, 27\% for GRPO-CARE, and 37\% for DRA-GRPO; the supplementary 4B audit shows the same direction relative to GRPO.
These findings show that agreement can be useful in the evaluated mathematical reasoning setting and that final-answer accuracy leaves important differences in reasoning validity unresolved.
A further question is how to combine the discovery of diverse solution paths with learning from reliable ones.

\section*{Limitations}
\label{sec:limitations}

\paragraph{Agreement as a proxy for reasoning validity.}
Multiple rollouts can share the same unresolved error and receive high agreement scores.
Conversely, a valid solution strategy represented by few rollouts may receive a low score.
Both valid low-agreement traces and spurious high-agreement traces occur in our paired analysis.
Appendix~\ref{app:agreement_counterexample} provides a paired example of a high-agreement spurious trace and a low-agreement valid trace.
The usefulness of agreement therefore depends on the reasoning patterns represented in the sampled group.

\paragraph{Scope of evaluation.}
Our experiments cover mathematical reasoning with 1.5B and 4B models trained on 7K examples.
The paired diagnostic uses 32 rollouts and compares agreement among verified-correct outputs; training uses the full on-policy group with six rollouts.
Our reward-construction ablations evaluate the reference-set choice, while the validity audits compare traces on problems answered correctly by all participating models.
\ours{} relies on reliable final-answer verification to determine which rollouts receive the agreement reward.
Further study should examine larger models, other reasoning domains and training budgets, and how the choice between diversity and agreement varies across policy distributions and stages of training.

\section*{Ethical Considerations}
This work uses publicly available English mathematical reasoning benchmarks and publicly released models, which contain no personally identifying or offensive content (Appendix~\ref{app:artifacts}).
A potential risk is that agreement-based rewards may reinforce reasoning patterns that are common but flawed, giving users unwarranted confidence in model-generated explanations; we therefore report spurious-reasoning rates alongside accuracy rather than treating rewarded traces as verified.

\bibliography{custom}

\appendix
\section{Data and Model Artifacts}
\label{app:artifacts}

Table~\ref{tab:artifact_details} summarizes the external data, code, and model artifacts used in our experiments.
All datasets are English mathematical or scientific reasoning problem sets.
They cover competition mathematics, formal mathematical problem solving, and quantitative scientific reasoning.
We use Open-RS as the training dataset, while the remaining datasets are used only for zero-shot evaluation.
We use DeepSeek-R1-Distill-Qwen-1.5B and Qwen3-4B as base models and implement training using the Open-RS codebase.

\begin{table*}[t]
\centering
\small
\setlength{\tabcolsep}{4pt}
\renewcommand{\arraystretch}{1.05}
\resizebox{\textwidth}{!}{
\begin{tabular}{llllr}
\toprule
\textbf{Artifact} & \textbf{Use} & \textbf{Repository} & \textbf{License / terms source} & \textbf{\# Examples} \\
\midrule
Open-RS dataset & Train & \href{https://huggingface.co/datasets/knoveleng/open-rs}{knoveleng/open-rs} & MIT & 7,000 \\
Open-RS codebase & Training pipeline & \href{https://github.com/knoveleng/open-rs}{knoveleng/open-rs} & MIT (official repo) & -- \\
AIME26 & Evaluation & \href{https://huggingface.co/datasets/math-ai/aime26}{math-ai/aime26} & Apache-2.0 & 30 \\
MATH-500 & Evaluation & \href{https://huggingface.co/datasets/HuggingFaceH4/MATH-500}{HuggingFaceH4/MATH-500} & MIT (source repo) & 500 \\
AMC23 & Evaluation & \href{https://huggingface.co/datasets/knoveleng/AMC-23}{knoveleng/AMC-23} & not specified & 40 \\
Minerva & Evaluation & \href{https://huggingface.co/datasets/knoveleng/Minerva-Math}{knoveleng/Minerva-Math} & MIT (equivalent mirror) & 272 \\
OlympiadBench & Evaluation & \href{https://huggingface.co/datasets/knoveleng/OlympiadBench}{knoveleng/OlympiadBench} & MIT (official repo) & 675 \\
DeepSeek-R1-Distill-Qwen-1.5B & Base model & \href{https://huggingface.co/deepseek-ai/DeepSeek-R1-Distill-Qwen-1.5B}{deepseek-ai/DeepSeek-R1-Distill-Qwen-1.5B} & MIT & -- \\
Qwen3-4B & Base model & \href{https://huggingface.co/Qwen/Qwen3-4B}{Qwen/Qwen3-4B} & Apache-2.0 & -- \\
DRA-GRPO & Evaluation checkpoint & \href{https://huggingface.co/SpiceRL/DRA-GRPO}{SpiceRL/DRA-GRPO} & CC-BY-4.0 (model card) & -- \\
all-MiniLM-L6-v2 & Semantic similarity & \href{https://huggingface.co/sentence-transformers/all-MiniLM-L6-v2}{sentence-transformers/all-MiniLM-L6-v2} & Apache-2.0 & -- \\
\bottomrule
\end{tabular}
}
\caption{
External data, code, and model artifacts used in this work.
The table reports each artifact's role in our experiments, repository, license or terms information available from the relevant repository or source, and dataset size when applicable.
}
\label{tab:artifact_details}
\end{table*}

\paragraph{Licenses and terms.}
The Open-RS dataset, Open-RS codebase, and DeepSeek-R1-Distill-Qwen-1.5B model are released under the MIT License, while AIME26 and Qwen3-4B are released under Apache-2.0 according to their Hugging Face cards or official repositories.
For MATH-500, the Hugging Face dataset card points to the \href{https://github.com/openai/prm800k}{OpenAI PRM800K source repository}, which is released under the MIT License.
For Minerva, the Hugging Face dataset card we used does not explicitly specify a license; however, we found an equivalent 272-example Hugging Face mirror, \href{https://huggingface.co/datasets/svc-huggingface/minerva-math}{svc-huggingface/minerva-math}, released under the MIT License.
For OlympiadBench, the official \href{https://github.com/OpenBMB/OlympiadBench}{OpenBMB repository} is released under the MIT License.
For AMC23, we did not find an explicit license listed on the current Hugging Face dataset card.
We use AMC23 only for academic evaluation.

\paragraph{Intended use.}
We use the Open-RS codebase to fine-tune the base models on Open-RS and evaluate them on the five benchmarks.
We do not redistribute modified versions of datasets with unclear licensing terms.

\section{Training Details}
\label{app:hyperparams}

Table~\ref{tab:hyperparams} summarizes the shared training hyperparameters used for the GRPO baseline, GRPO-CARE, and GRPO-\ours{} unless otherwise specified.
We compute semantic similarity using embeddings from \texttt{sentence-transformers/all-MiniLM-L6-v2} and lexical overlap using token-level F1.
\paragraph{GRPO-CARE reimplementation.}
Our GRPO-CARE baseline uses the shared settings in Table~\ref{tab:hyperparams}, with CARE enabled using base reward index 0, minimum correct reward 0.1, bonus coefficient 0.5, confidence upper bound 0.95, consistency margin 0.01, reference EMA decay 0.995, and EMA update interval 10.
Answers are extracted from the \texttt{<answer>}...\texttt{</answer>} span.
\begin{table}[t]
\centering
\footnotesize
\setlength{\tabcolsep}{4pt}
\renewcommand{\arraystretch}{1.05}
\begin{tabular}{p{0.42\linewidth}p{0.46\linewidth}}
\toprule
\textbf{Parameter} & \textbf{Value} \\
\midrule

\multicolumn{2}{l}{\textbf{\textit{General Settings}}} \\
Precision & bf16 \\
vLLM & enabled, colocate mode \\
vLLM memory ratio & 0.1 \\
vLLM max length & 4608 \\
Evaluation during training & false \\

\midrule
\multicolumn{2}{l}{\textbf{\textit{Training Configuration}}} \\
Gradient accumulation & 8 \\
Gradient checkpointing & true \\
Learning rate & $1.0\times10^{-6}$ \\
Scheduler & cosine with min-lr ratio 0.1 \\
Warmup ratio & 0.1 \\
Max steps / epochs & 500 / 1 \\
Train batch size & 12 per device \\

\midrule
\multicolumn{2}{l}{\textbf{\textit{Generation Settings}}} \\
Max completion length & 3584 \\
Number of generations & 6 \\
Temperature & 0.7 \\
Default training seed & 2025 \\

\midrule
\multicolumn{2}{l}{\textbf{\textit{Reward Configuration}}} \\
Reward functions & $r_{\text{fmt}}$, $r_{\text{acc}}$, $r_{\text{\ours{}}}$ \\
Reward weights & 1.0, 2.0, 1.0 \\
Agreement reference & all rollouts \\
Reward target & verified-correct rollouts only \\
Mixing coefficient $\alpha$ & 0.5 \\

\bottomrule
\end{tabular}
\caption{
Hyperparameter setup for GRPO training.
The multi-seed experiments use three training seeds.
For \ours{}, the default configuration uses the full rollout group as the agreement reference and assigns the rollout-agreement reward only to verified-correct rollouts.
}
\label{tab:hyperparams}
\end{table}

\paragraph{GRPO w/ LLM reward reimplementation.}
The 1.5B baseline uses the same 7K Open-RS training examples and 500 training steps as the main experiment, with six rollouts sampled per prompt.
We use \texttt{gpt-5-mini} with minimal reasoning effort to score the trajectories in each rollout group.
The judge returns one score per trajectory, which supplies the RULER reward in GRPO.
We evaluate the resulting model with the same pass@1 protocol as the other models trained in our environment.

\paragraph{Motivating-analysis generation settings.}
\label{app:motivation_generation}
For the analysis in Section~\ref{sec:motivation}, we sample 32 rollouts per problem on the 272-problem Minerva benchmark using temperature 0.6, top-$p$ 0.95, and a maximum generation length of 32,768 tokens.
We compute agreement among verified-correct rollouts with a semantic--lexical mixing coefficient of $\alpha=0.5$.
We retain problems with at least five verified-correct rollouts and pair the highest- and lowest-agreement traces, yielding 126 pairs.

\section{Resource-Overhead Check}
\label{app:resource_overhead}

Table~\ref{tab:resource_overhead} compares the resource use of outcome-only GRPO and GRPO-\ours{} under matched training settings.
Both runs complete 500 training steps on 8 NVIDIA B200 GPUs.
The values are taken from W\&B run summaries, logged profiler metrics, and sampled system telemetry.
The two runs have comparable wall-clock times.

\begin{table*}[t]
\centering
\small
\setlength{\tabcolsep}{3pt}
\renewcommand{\arraystretch}{1.05}
\resizebox{\linewidth}{!}{
\begin{tabular}{lrrrrrr}
\toprule
\textbf{Method} &
\textbf{Runtime} &
\textbf{GPU-hours} &
\textbf{Step time} &
\textbf{Peak memory} &
\textbf{Reward comp.} &
\textbf{\ours{} comp.} \\
&
\textbf{(h)} &
&
\textbf{(s)} &
\textbf{(GiB/GPU)} &
\textbf{(ms)} &
\textbf{(ms)} \\
\midrule
GRPO & 11.34 & 90.75 & 80.67 & 178.3 & 27.3 & -- \\
GRPO-\ours{} & 11.27 & 90.14 & 81.35 & 178.3 & 47.7 & 23.6 \\
\bottomrule
\end{tabular}
}
\caption{
Resource-overhead check for outcome-only GRPO and GRPO-\ours{} under matched training runs.
GPU-hours are computed from wall-clock time and 8 GPUs.
Reward-computation times are averages from logged profiler metrics.
}
\label{tab:resource_overhead}
\end{table*}

\section{Bootstrap Confidence Intervals}
\label{app:bootstrap_ci}

Table~\ref{tab:bootstrap_ci} reports paired bootstrap confidence intervals over evaluation problems for the comparisons in Section~\ref{sec:main_results}.
These intervals estimate uncertainty from the finite evaluation set.
All intervals are computed from paired per-problem outcomes in our evaluations.
For DRA-GRPO, the interval corresponds to our pass@4 evaluation of its released checkpoint; its pass@1 row in Table~\ref{tab:main_table} combines published scores on four benchmarks with our AIME26 evaluation.

\begin{table}[t]
\centering
\small
\setlength{\tabcolsep}{4.5pt}
\renewcommand{\arraystretch}{1.05}
\begin{tabular}{lccc}
\toprule
\textbf{Baseline} & \textbf{Metric} & \textbf{$\Delta$} & \textbf{95\% CI} \\
\midrule
GRPO & Pass@1 & +6.0 & [1.6, 10.6] \\
GRPO-CARE & Pass@1 & +5.5 & [0.9, 10.0] \\
GRPO & Pass@4 & +2.1 & [-1.2, 5.6] \\
GRPO-CARE & Pass@4 & +3.2 & [0.7, 6.2] \\
DRA-GRPO & Pass@4 & +4.7 & [1.5, 8.2] \\
\bottomrule
\end{tabular}
\caption{
Paired bootstrap confidence intervals over evaluation problems.
Deltas compare GRPO-\ours{} against each baseline and are macro-average percentage-point differences across the five evaluation benchmarks.
}
\label{tab:bootstrap_ci}
\end{table}

\section{Results Across Training Seeds}
\label{app:seed_results}

Table~\ref{tab:rebuttal_seeds} reports the five-benchmark mean pass@1 for each training seed at both model sizes.
The per-benchmark means and standard deviations across seeds are reported in Table~\ref{tab:rebuttal_performance}.

\begin{table*}[t]
\centering
\small
\setlength{\tabcolsep}{5pt}
\renewcommand{\arraystretch}{1.05}
\begin{tabular}{lrrrrrr}
\toprule
& \multicolumn{3}{c}{\textbf{1.5B}} & \multicolumn{3}{c}{\textbf{4B}} \\
\cmidrule(lr){2-4}\cmidrule(lr){5-7}
\textbf{Training seed} & \textbf{GRPO} & \textbf{GRPO-\ours{}} & \textbf{$\Delta$} & \textbf{GRPO} & \textbf{GRPO-\ours{}} & \textbf{$\Delta$} \\
\midrule
0 & 53.21 & \textbf{53.31} & +0.10 & \textbf{75.41} & 74.35 & $-1.06$ \\
7 & \textbf{52.19} & 51.27 & $-0.92$ & 73.16 & \textbf{75.93} & +2.77 \\
2025 & 49.50 & \textbf{55.52} & +6.02 & 75.52 & \textbf{78.15} & +2.64 \\
\bottomrule
\end{tabular}
\caption{Five-benchmark mean pass@1 (\%) for individual training runs. Each score is the macro-average over AIME26, MATH-500, AMC23, Minerva, and OlympiadBench. $\Delta$ is the GRPO-\ours{} improvement in percentage points, computed from unrounded scores. Bold indicates the higher score within each model size and training seed.}
\label{tab:rebuttal_seeds}
\end{table*}

\section{Detailed Results for Semantic--Lexical Mixing Ablation}
\label{app:alpha_ablation}

Table~\ref{tab:alpha_ablation} reports the per-benchmark results for the semantic--lexical mixing ablation discussed in Section~\ref{sec:agreement_sensitivity}.
We use the full on-policy rollout group as the reference set and assign the agreement reward only to verified-correct rollouts, as in the default \ours{} configuration.
We vary only the mixing coefficient $\alpha$ in Eq.~\ref{eq:s_mix}.


\begin{table*}[t]
\centering
\small
\setlength{\tabcolsep}{5.0pt}
\renewcommand{\arraystretch}{1.05}
\begin{tabular}{lcccccc}
\toprule
\textbf{$\alpha$} & \textbf{AIME26} & \textbf{MATH-500} & \textbf{AMC23} & \textbf{Minerva} & \textbf{OlympiadBench} & \textbf{Avg.} \\
\midrule
0.00 \; (Lexical only) & 23.3 & 84.0 & 70.0 & 34.2 & 52.7 & 52.8 \\
0.25 & 26.7 & 85.2 & 77.5 & 30.9 & 50.7 & 54.2 \\
\rowcolor{gray!15}
0.50 & 26.7 & 84.8 & 80.0 & 30.9 & 55.3 & \textbf{55.5} \\
0.75 & 30.0 & 85.2 & 72.5 & 30.9 & 52.9 & 54.3 \\
1.00 \; (Semantic only) & 23.3 & 83.6 & 70.0 & 32.7 & 50.5 & 52.0 \\
\bottomrule
\end{tabular}
\caption{
Per-benchmark ablation of the semantic--lexical mixing coefficient $\alpha$ in the mixed similarity score.
Numbers are in \%.
$\alpha=0$ denotes lexical-only agreement, and $\alpha=1$ denotes semantic-only agreement.
}
\label{tab:alpha_ablation}
\end{table*}

\section{Detailed Results for Rollout-Agreement Weight Ablation}
\label{app:weight_ablation}

Table~\ref{tab:weight_ablation} reports the per-benchmark results for the rollout-agreement reward weight ablation discussed in Section~\ref{sec:agreement_sensitivity}.
We use the full on-policy rollout group as the reference set and assign the agreement reward only to verified-correct rollouts.
We vary only the reward weight $w_{\ours}$.

\begin{table*}[t]
\centering
\small
\setlength{\tabcolsep}{5.0pt}
\renewcommand{\arraystretch}{1.05}
\begin{tabular}{lcccccc}
\toprule
\textbf{$w_{\ours}$} & \textbf{AIME26} & \textbf{MATH-500} & \textbf{AMC23} & \textbf{Minerva} & \textbf{OlympiadBench} & \textbf{Avg.} \\
\midrule
2.0 & \textbf{30.0} & 82.8 & 72.5 & \textbf{32.7} & 50.4 & 53.7 \\
1.5 & \textbf{30.0} & 82.8 & \textbf{80.0} & 30.9 & 54.2 & \textbf{55.6} \\
\rowcolor{gray!15}
1.0 & 26.7 & \textbf{84.8} & \textbf{80.0} & 30.9 & \textbf{55.3} & 55.5 \\
0.5 & 20.0 & 83.2 & 72.5 & 30.9 & 52.6 & 51.8 \\
\bottomrule
\end{tabular}
\caption{
Per-benchmark ablation of the rollout-agreement reward weight $w_{\ours}$.
Numbers are in \%.
The shaded row denotes the default setting used in the main experiments.
}
\label{tab:weight_ablation}
\end{table*}

\section{Consensus-Conditioned Evaluation Details}
\label{app:consensus_conditioned}

Table~\ref{tab:consensus_conditioned_full} reports per-benchmark results for the consensus-conditioned evaluation in Section~\ref{sec:consensus_conditioned}.
As in the main analysis, strong consensus denotes problems where at least three of four sampled rollouts produce the same extracted final answer, and weak-consensus metrics are computed over the remaining problems.

\begin{table*}[t]
\centering
\small
\setlength{\tabcolsep}{5.5pt}
\renewcommand{\arraystretch}{1.05}
\begin{tabular}{llccc}
\toprule
\textbf{Dataset} & \textbf{Model} & \textbf{Strong Cons.} & \textbf{Weak Pass@1} & \textbf{Weak Pass@4} \\
\midrule
\multirow{4}{*}{AIME26}
& GRPO & 13.3 & 3.8 & 26.9 \\
& GRPO-CARE & 23.3 & 8.7 & 17.4 \\
& DRA-GRPO & 13.3 & 11.5 & 15.4 \\
& GRPO-\ours{} & 20.0 & \textbf{16.7} & \textbf{33.3} \\
\midrule
\multirow{4}{*}{MATH-500}
& GRPO & 82.6 & 32.2 & 66.7 \\
& GRPO-CARE & 82.0 & 23.3 & 63.3 \\
& DRA-GRPO & 84.4 & 32.1 & 65.4 \\
& GRPO-\ours{} & 82.6 & \textbf{36.8} & \textbf{67.8} \\
\midrule
\multirow{4}{*}{AMC23}
& GRPO & 72.5 & 27.3 & 63.6 \\
& GRPO-CARE & 80.0 & 0.0 & 50.0 \\
& DRA-GRPO & 70.0 & 16.7 & 75.0 \\
& GRPO-\ours{} & 67.5 & \textbf{38.5} & \textbf{76.9} \\
\midrule
\multirow{4}{*}{Minerva}
& GRPO & 26.1 & 15.9 & 33.3 \\
& GRPO-CARE & 29.4 & 13.5 & 33.3 \\
& DRA-GRPO & 25.7 & 18.8 & 28.7 \\
& GRPO-\ours{} & 23.5 & \textbf{19.7} & \textbf{37.0} \\
\midrule
\multirow{4}{*}{OlympiadBench}
& GRPO & 49.0 & 26.7 & 49.1 \\
& GRPO-CARE & 48.0 & 28.2 & \textbf{51.0} \\
& DRA-GRPO & 48.6 & 25.6 & 49.9 \\
& GRPO-\ours{} & 50.4 & \textbf{28.7} & 46.9 \\
\midrule
\multirow{4}{*}{MacroAvg}
& GRPO & 48.7 & 21.2 & 47.9 \\
& GRPO-CARE & 52.5 & 14.8 & 43.0 \\
& DRA-GRPO & 48.4 & 20.9 & 46.9 \\
& GRPO-\ours{} & 48.8 & \textbf{28.1} & \textbf{52.4} \\
\bottomrule
\end{tabular}
\caption{
Per-benchmark consensus-conditioned evaluation over four sampled rollouts.
Strong consensus denotes the fraction of problems where at least three rollouts produce the same extracted final answer.
Weak Pass@1 and Weak Pass@4 are computed only over the remaining weak-consensus problems.
Numbers are in \%. DRA-GRPO is evaluated in our environment using its released checkpoint.
}
\label{tab:consensus_conditioned_full}
\end{table*}

\section{Supplementary Qwen3-4B Human Analysis}
\label{app:human_4b_qual}

We examine reasoning validity at the 4B scale using GRPO and GRPO-\ours{} Qwen3-4B models trained under matched settings.
We sample 100 problems where both models produce a correct first rollout.
The traces are annotated with model identities hidden, using the guideline in Appendix~\ref{app:qual_prompt}.
Table~\ref{tab:human_4b_validity} shows that the valid-trace rate increases from 49\% to 56\%, while the spurious-trace rate decreases from 36\% to 29\%; the self-corrected rate remains 15\%.

\begin{table}[t]
\centering
\small
\setlength{\tabcolsep}{4pt}
\begin{tabular}{lccc}
\toprule
\textbf{Model} & \textbf{Valid ($\uparrow$)} & \textbf{Self-corrected ($-$)} & \textbf{Spurious ($\downarrow$)} \\
\midrule
GRPO & 49 & 15 & 36 \\
GRPO-\ours{} & \textbf{56} & 15 & \textbf{29} \\
\bottomrule
\end{tabular}
\caption{Supplementary blind human evaluation of the Qwen3-4B models trained under matched settings on 100 sampled problems where both models produce a correct first rollout. Entries are trace counts; arrows indicate the preferred direction.}
\label{tab:human_4b_validity}
\end{table}

\section{LLM-Based Robustness Check for the 1.5B Analysis}
\label{app:llm_qual}

We apply GPT-6.1 Sol to the traces from the same 100 shared correct-answer problems using the annotation guideline in Appendix~\ref{app:qual_prompt}.
The judge receives the problem, target answer, and candidate reasoning, without the human label.
We use low reasoning effort and a maximum output length of 8192 tokens; no temperature or top-$p$ parameter is specified.
Human--LLM inter-annotator agreement (IAA), measured as exact agreement across the three categories, is 79.25\%, with Cohen's $\kappa=0.665$.

Table~\ref{tab:llm_validity} reports the label percentages for all four models.
The judge also assigns GRPO-\ours{} the lowest spurious rate, while assigning higher spurious rates to all four models than the human annotator.
Human annotations remain the primary evaluation, with the LLM labels providing a complementary check.
Table~\ref{tab:llm_human_confusion} provides the full confusion matrix.

\begin{table}[t]
\centering
\small
\setlength{\tabcolsep}{3pt}
\begin{tabular}{lccc}
\toprule
\textbf{Model} & \textbf{Valid ($\uparrow$)} & \textbf{Self-corr. ($-$)} & \textbf{Spurious ($\downarrow$)} \\
\midrule
GRPO & 45 & 13 & 42 \\
GRPO-\ours{} & 47 & 18 & \textbf{35} \\
GRPO-CARE & \textbf{48} & 15 & 37 \\
DRA-GRPO & 42 & 18 & 40 \\
\bottomrule
\end{tabular}
\caption{GPT-6.1 Sol (low reasoning effort) reasoning-validity labels for all four 1.5B models on the same 100 shared correct-answer problems. Entries are percentages; arrows indicate the preferred direction.}
\label{tab:llm_validity}
\end{table}

\begin{table}[t]
\centering
\small
\begin{tabular}{lrrr}
\toprule
\textbf{Human $\backslash$ LLM} & \textbf{Valid} & \textbf{Self-corr.} & \textbf{Spurious} \\
\midrule
Valid & 177 & 12 & 23 \\
Self-corrected & 1 & 41 & 32 \\
Spurious & 4 & 11 & 99 \\
\bottomrule
\end{tabular}
\caption{Human--LLM label agreement on the shared correct-answer traces. Rows are human labels; columns are GPT-6.1 Sol labels.}
\label{tab:llm_human_confusion}
\end{table}

\section{Case Study}
\label{app:case_study}

Figure~\ref{fig:qualitative_case_study} compares three responses to the same optimization problem.
All three reach the correct final answer, \(\boxed{64}\), but differ in reasoning validity.

In the GRPO response, the model initially derives the correct solution but later introduces an erroneous re-check. 
In particular, it incorrectly derives \(c=a/2\) and \(d=a/4\), obtains the candidate value \(203.5\), and then states that this value is ``way less than 64.'' 
Although the final answer returns to \(64\), this numerical comparison error is not explicitly corrected, so we label the trace as spurious.

The GRPO-\ours{} response applies the Lagrange multiplier conditions to obtain \(a=b\), \(c=2a\), and \(d=4a\).
It substitutes these relations into the constraint and verifies the minimum using the Cauchy-Schwarz inequality. 
This response is therefore labeled valid.

The GRPO-CARE response initially derives incorrect variable relations, then revisits the derivation after checking the Cauchy-Schwarz lower bound.
It re-examines the Lagrange equations, corrects the variable relations, and derives the minimum.
We label this response as self-corrected.

\newpage
\begin{figure*}[t]
\centering
\scriptsize
\setlength{\abovedisplayskip}{2pt}
\setlength{\belowdisplayskip}{2pt}

\begin{tcolorbox}[
  enhanced,
  colback=casebg,
  colframe=caseframe,
  colbacktitle=casebar,
  coltitle=casetitle,
  boxrule=0.5pt,
  arc=1.5mm,
  left=1.8mm,
  right=1.8mm,
  top=1mm,
  bottom=1mm,
  fontupper=\scriptsize,
  title={\textbf{Case Study: Same Correct Answer, Different Reasoning Validity}},
  fonttitle=\small\bfseries
]
\textbf{Problem.}
Find the minimum of
\[
\frac{1}{a}+\frac{1}{b}+\frac{4}{c}+\frac{16}{d}
\quad
\text{s.t.}
\quad
a+b+c+d=1,\quad a,b,c,d>0.
\]
\textbf{Target answer:} \(\boxed{64}\).
\end{tcolorbox}

\vspace{0.6mm}

\begin{caseblock}{GRPO}{\textcolor{errfg}{Spurious}}

\textbf{Initial correct derivation.}
\emph{
``From equations 1 and 3:
\(-1/a^2=-4/c^2\Rightarrow 1/a^2=4/c^2\Rightarrow
\goodm{c=2a}\).
From equations 1 and 4:
\(-1/a^2=-16/d^2\Rightarrow
\goodm{d=4a}\).
So \(a+a+2a+4a=8a=1\), and the expression evaluates to
\(\goodm{64}\).''
}

\vspace{0.35em}

\textbf{Later erroneous re-check.}
\emph{
``Taking square roots, we get
\(\badm{a=b=2c=4d}\).
Wait, hold on ... from \(4c^2=a^2\), so
\(\badm{c=a/2}\), and from \(16d^2=a^2\), so
\(\badm{d=a/4}\).''
}

\vspace{0.35em}

\textbf{Unresolved numerical comparison.}
\emph{
``Now, plug these back into the original expression:
\(1/a=2.75\), \(1/b=2.75\), \(4/c=22\), \(16/d=176\).
Adding them up gives \(\badm{203.5}\).
Wait, that's \badspan{way less than 64}.
So that suggests that the minimum is \(\badm{203.5}\).''
}

\vspace{0.35em}

\textbf{Final output.}
\emph{
``Therefore, the correct solution is \(a=1/8\), \(b=1/8\), \(c=1/4\), \(d=1/2\), and the expression evaluates to
\(\goodm{64}\).''
}

\end{caseblock}

\vspace{0.6mm}

\begin{caseblock}{GRPO-\ours{}}{\textcolor{goodfg}{Valid}}

\textbf{Lagrange setup.}
\emph{
``Let's set up the function to minimize:
\(f(a,b,c,d)=1/a+1/b+4/c+16/d\).
The constraint is \(g(a,b,c,d)=a+b+c+d-1=0\).
The method of Lagrange multipliers says that
\(\nabla f=\lambda \nabla g\).''
}

\vspace{0.3em}

\textbf{Correct stationarity conditions.}
\emph{
``Computing the gradients gives
\((-1/a^2,-1/b^2,-4/c^2,-16/d^2)\).
Setting them proportional:
\[
\goodm{
-\frac{1}{a^2}=\lambda,\quad
-\frac{1}{b^2}=\lambda,\quad
-\frac{4}{c^2}=\lambda,\quad
-\frac{16}{d^2}=\lambda
}
\]
From equations 1 and 2, \(a=b\).
From equations 1 and 3, \(c^2=4a^2\), so \(\goodm{c=2a}\).
From equations 1 and 4, \(d^2=16a^2\), so \(\goodm{d=4a}\).''
}

\vspace{0.3em}

\textbf{Constraint substitution.}
\emph{
``So, we have \(\goodm{a=b,\ c=2a,\ d=4a}\).
Now substituting back into the constraint:
\[
\goodm{a+b+c+d=a+a+2a+4a=8a=1}
\]
Therefore, \(\goodm{a=1/8}\), and hence
\[
\goodm{
a=b=\frac18,\quad c=\frac14,\quad d=\frac12
}.
\]''
}

\vspace{0.3em}

\textbf{Value and verification.}
\emph{
``Plugging these back into the original expression gives
\[
\goodm{
\frac{1}{1/8}+\frac{1}{1/8}+\frac{4}{1/4}+\frac{16}{1/2}
=8+8+16+32=64
}.
\]
Wait, that seems like a big number. Let me double-check the substitution.
\(4/c=16\) and \(16/d=32\), so \(8+8+16+32=64\).''
}

\vspace{0.3em}

\textbf{Independent lower-bound check.}
\emph{
``Using Cauchy-Schwarz,
\[
\goodm{
\left(
\frac1a+\frac1b+\frac4c+\frac{16}{d}
\right)(a+b+c+d)
\geq
(1+1+2+4)^2=64
}.
\]
Since \(a+b+c+d=1\), the expression is at least \(\goodm{64}\).
This is consistent with the Lagrange multiplier result.''
}

\vspace{0.2em}

\textbf{Final output.}
\emph{
``Therefore, the minimum value is \(\goodm{\boxed{64}}\).''
}

\end{caseblock}

\vspace{0.6mm}
\begin{caseblock}{GRPO-CARE}{\textcolor{fixfg}{Self-corrected}}

\textbf{Initial wrong path.}
\emph{
``Therefore, \(a^2=b^2=4c^2=16d^2\). Since all variables are positive,
\(\badm{a=b=2c=4d}\).
So, let me express \(a\), \(b\), and \(d\) in terms of \(c\):
\(\badm{a=2c,\ b=2c,\ d=c/4}\).
Substituting into the constraint gives \(c=4/21\), \(a=b=8/21\), and \(d=1/21\).''
}

\smallskip

\emph{
``Adding them up gives
\(\badm{21/8+21/8+21+336=1449/4=362.25}\).
So, \(\badm{1449/4}\) is the minimal value.''
}

\smallskip

\textbf{Lower-bound check.}
\emph{
``Now, using Cauchy-Schwarz,
\[
\fixm{
\left(
\frac{1^2}{a}+\frac{1^2}{b}+\frac{2^2}{c}+\frac{4^2}{d}
\right)(a+b+c+d)
\geq
(1+1+2+4)^2=64
}
\]
Since \(a+b+c+d=1\), this gives
\(\fixm{S\geq 64}\).''
}

\smallskip

\textbf{Self-correction.}
\emph{
``The Cauchy-Schwarz inequality gives a lower bound of \(64\), but using Lagrange multipliers I got a higher value.
This suggests that perhaps the minimal value is \(64\), but my Lagrange multipliers gave a higher value, which is a contradiction.
So, \(\fixspan{I must have made a mistake in applying Lagrange multipliers}\).''
}

\smallskip

\textbf{Corrected Lagrange path.}
\emph{
``Let me re-examine the steps:
\(-1/a^2=-1/b^2\Rightarrow a=b\),
\(-1/a^2=-4/c^2\Rightarrow \goodm{c=2a}\),
and
\(-1/a^2=-16/d^2\Rightarrow \goodm{d=4a}\).
Thus,
\(\goodm{a=b,\ c=2a,\ d=4a}\).''
}

\smallskip

\textbf{Final output.}
\emph{
``Then \(a+a+2a+4a=8a=1\), so
\(\goodm{a=b=1/8,\ c=1/4,\ d=1/2}\).
The expression is
\(\goodm{8+8+16+32=64}\).
Therefore, the minimal value is \(\goodm{\boxed{64}}\).''
}

\end{caseblock}

\caption{
Qualitative case study comparing reasoning validity across methods.
All responses reach the same correct final answer, but exhibit different reasoning traces:
spurious, valid, and self-corrected.
}
\label{fig:qualitative_case_study}
\end{figure*}

\subsection{A Counterexample to Agreement-Based Ranking}
\label{app:agreement_counterexample}

A pair from the GRPO-trained model in Section~\ref{sec:motivation} illustrates how agreement can rank a spurious trace above a valid one.
The problem asks for the free-trade price of barley in a small country with demand \(Q^D=4-P\), supply \(Q^S=P\), and a world price of \(\$1\) per bushel.
Both traces return the correct price, \(\$1\).
At this price, demand is three bushels and supply is one, so the country imports two bushels.
The higher-agreement trace repeatedly describes this shortage as exports, while the lower-agreement trace correctly identifies imports (Table~\ref{tab:agreement_counterexample}).
The export error remains unresolved, giving the higher-agreement trace its spurious label in the blind annotation.

\begin{table*}[t]
\centering
\small
\renewcommand{\arraystretch}{1.15}
\begin{tabular}{@{}p{0.47\textwidth}p{0.47\textwidth}@{}}
\toprule
\textbf{Higher agreement: 0.788 (Spurious)} & \textbf{Lower agreement: 0.671 (Valid)} \\
\midrule
``If the world price is \$1, and the country's supply is Qs = P, then at P = 1, Qs = 1. The demand is Qd = 4 - 1 = 3. So, the country will export the difference, which is 2 bushels.''
&
``The world price is 1. So the demand in Moldavia is 3 and supply is 1, so 2 bushels will be imported. The price is still 1 because the world price is lower than the local equilibrium.'' \\
\midrule
\textbf{Verified final answer:} \(\$1\) & \textbf{Verified final answer:} \(\$1\) \\
\bottomrule
\end{tabular}
\caption{A paired counterexample from the 126-pair GRPO diagnostic. Excerpts are taken from the reasoning traces. Agreement scores use the other verified-correct rollouts in the same 32-rollout group, following Section~\ref{sec:motivation}; labels are from the original blind annotation. The higher-agreement trace retains an incorrect trade direction despite reaching the correct final price.}
\label{tab:agreement_counterexample}
\end{table*}

\section{Annotation Guideline for Reasoning Validity}
\label{app:qual_prompt}

We use the following annotation guideline for both human annotation and the 1.5B LLM robustness check.
The target final answer is assumed to be correct, and annotators judge only whether the reasoning process supports that answer.
The decision rules below are shared by human annotation and LLM evaluation; the JSON output instructions apply to the LLM judge.
The LLM judge settings are reported in Appendix~\ref{app:llm_qual}.

\clearpage

\begin{promptbox}[breakable]{Annotation guideline for reasoning-validity}
You are an expert auditor for mathematical reasoning-process validity.
Evaluate exactly ONE candidate reasoning process for ONE math problem.

The target final answer is assumed to be correct.
Do NOT judge final-answer accuracy.
Judge only whether the candidate reasoning process supports the target final answer.

You will receive:
- Problem
- Target final answer
- Candidate reasoning

Output only a parseable JSON object with exactly these keys:
{
"validity": "valid" | "self-corrected" | "spurious",
"error_type": ["none" | "wrong_evidence_or_condition" | "wrong_operation_or_formula" | "unit_or_scale_error" | "unsupported_step" | "internal_inconsistency" | "answer_rationale_mismatch" | "insufficient_reasoning" | "unnecessary_wrong_information" | "other"]
}

No explanation.
No markdown.
No extra keys.

Core decision rule:

Scan the entire candidate reasoning, including exploratory work, intermediate derivations, checks, alternative approaches, and final summary.
Do not inspect only the final answer line or only the final adopted derivation.
First decide whether the reasoning contains an actual error.
An actual error must be grounded in a concrete false mathematical, computational, definitional, condition-based, unit/scale, dimensional, or formal-representation claim in the text.
Do not infer an actual error merely from self-checking language such as "wait", "maybe", "let me check", "that's not right", hesitation, repeated recomputation, uncertainty, or ordinary exploration, unless the surrounding reasoning actually states a false value, formula, condition, definition, dimension, mapping, or calculation.
A stated incorrect arithmetic result, numerical equality, formula, condition, variable definition, dimensional relation, mapping, or formal representation counts as a concrete false claim when it is presented as part of the reasoning rather than only as a tentative hypothesis being tested.
If a concrete false claim is asserted as a result, premise, calculation, condition, relation, formula, definition, mapping, or justification, do not label the reasoning "valid".
A later correct answer, correct formula, or correct final path is not by itself an observable correction of an earlier false claim.
The earlier false claim must be directly addressed: it must be explicitly rejected, directly recomputed, replaced by the corresponding correct claim, or clearly abandoned without remaining active in the final adopted justification or final summary.
If a concrete false claim remains active in the final adopted justification or final summary, label "spurious" even if the final answer is correct.
If the problem asks for a formal representation, model, equation, or formulation, every equation, relation, variable definition, mapping, dimensional relation, and structural component that the reasoning introduces for that formulation must be checked for internal consistency and consistency with the problem requirements.
A formal relation is a concrete claim even if it is not repeated in the final answer line.
If the reasoning defines a quantity as scalar, vector, matrix, or structured object, then any later equation, mapping, or relation involving that quantity must be dimensionally and structurally compatible with that definition.
A structurally or dimensionally incompatible formal relation is a concrete false formal-representation claim, even if the final scalar equation or final numerical answer is correct.

Exploration vs. actual error:

A tentative exploratory route, hypothesis, candidate model, or alternative approach is not automatically an actual error if it is clearly being tested, compared, rejected, or abandoned before the final adopted path.
Do not penalize ordinary exploration merely because the reasoning tries an approach that does not work, checks a candidate, compares alternatives, or computes that a tentative route fails.
However, if the reasoning presents a concrete false result as part of the main reasoning path, treats a wrong route as a real working path before repairing it, or uses a wrong route to support the final answer, then it counts as an actual error.
A concrete false claim in the main adopted path, in a required formal representation, or in the final summary counts as an actual error.
A wrong claim is not observably corrected merely because it disappears from the final answer line or final summary.
The text must reject, fix, withdraw, replace, or clearly abandon the wrong claim or the route containing it.
A route-level correction is allowed.
If the reasoning explicitly finds that a route fails a stated condition, consistency check, substitution check, calculation check, or other problem-relevant check, then abandons that route and replaces it with an independent, stable, mathematically sound derivation, label "self-corrected" rather than "spurious", unless the wrong claim remains active or reappears in the final summary.

Use "valid" when:
- the reasoning contains enough support for the target final answer;
- the final adopted reasoning path is mathematically sound;
- there is no concrete false claim in any asserted reasoning step;
- any tentative hypothesis, candidate model, or exploratory suggestion is explicitly tentative and is not used as a result, premise, or final justification;
- there is no unresolved or incompatible claim in the main adopted path, required formal representation, or final summary.

Harmless self-checking, hesitation, verbosity, repeated verification, tentative exploration, or minor wording imprecision is allowed only when it does not assert a concrete false mathematical, computational, definitional, condition-based, unit/scale, dimensional, or formal-representation claim.

Use "self-corrected" when:
- the reasoning contains an actual intermediate error; and
- that error, or the route containing that error, is later observably corrected, rejected, withdrawn, abandoned, or replaced; and
- after the correction, the final adopted path is sound, stable, and supports the target final answer.

An actual error in a check, endpoint test, sanity check, intermediate derivation, or final summary can count as an intermediate error if it asserts a concrete false claim.
If such an actual error is observably fixed and the final path is sound, label "self-corrected", not "valid".

Use "spurious" when:
- the reasoning is too vague or insufficient to verify support;
- the target answer does not follow from the stated reasoning;
- an actual error remains unresolved;
- an earlier wrong claim is silently bypassed rather than corrected or abandoned;
- a wrong claim merely disappears from the final answer without being rejected, fixed, withdrawn, replaced, or clearly abandoned;
- incompatible formulas, values, definitions, dimensions, mappings, conditions, or reasoning paths remain active or appear in the final summary;
- a required formal representation contains an unresolved structural, dimensional, definitional, mapping, or representation error;
- the final adopted path is unclear or unstable.

A final adopted path is stable only if the reasoning clearly settles on one compatible derivation, one compatible interpretation of the problem conditions, and one compatible final justification.

Error types:

- none:
Use only with validity="valid".

- wrong_evidence_or_condition:
The reasoning misreads or misuses a problem condition, value, premise, constraint, domain, case, definition, diagram/table information, or previous context.

- wrong_operation_or_formula:
The reasoning uses a wrong formula, equation setup, algebraic manipulation, arithmetic procedure, sign, derivative, integral, inequality transformation, comparison, calculation, dimensional operation, mapping, or formal representation.

- unit_or_scale_error:
The reasoning makes a unit, scale, exponent, percentage, decimal, dimensional, radians/degrees, normalization, or order-of-magnitude error.

- unsupported_step:
The reasoning asserts an essential conclusion without enough support, or effectively guesses/rationalizes the target answer.

- internal_inconsistency:
The reasoning contains unresolved conflicting values, equations, definitions, dimensions, mappings, signs, cases, or intermediate conclusions.

- answer_rationale_mismatch:
The reasoning supports a different answer from the target final answer, or its stated conclusion does not match the target final answer.

- insufficient_reasoning:
The reasoning is empty, nearly empty, or too thin to verify that the target final answer follows.

- unnecessary_wrong_information:
The reasoning includes a clearly false side claim that is not needed for the solution and is not better captured by a more specific error type.

- other:
Use only when a real self-correction or spurious issue does not fit the listed categories.

Error type policy:
- valid -> ["none"]
- self-corrected/spurious -> never use ["none"].
- Use the minimal sufficient set of error types.
- Usually choose 1 or 2 representative error types.
- Prefer the root error over downstream consequences.
- For self-corrected, tag the corrected error type.
- For spurious, tag the unresolved core error type.

Tie-breaks:
- If there is no actual error and the reasoning is sufficient, choose "valid".
- If the only evidence of an error is self-checking language, hesitation, repeated recomputation, or tentative exploration without a concrete false claim, do not treat it as an actual error.
- If a tentative exploratory route or alternative model is clearly checked and abandoned before the final adopted path, do not count it as an actual error by itself.
- If a stated arithmetic result, numerical equality, formula, condition, definition, dimensional relation, mapping, or formal representation is concretely false and is part of the reasoning rather than only a tentative hypothesis, count it as an actual error.
- If there is an actual error that is observably corrected, rejected, withdrawn, abandoned, or replaced, and the final path is sound, choose "self-corrected".
- If a failed route is explicitly rejected because it fails a stated condition or consistency check, and the reasoning replaces it with an independent sound route, choose "self-corrected" rather than "spurious".
- If there is an actual error that is not observably corrected or abandoned, choose "spurious".
- If it is unclear whether an actual error was corrected or abandoned, choose "spurious".
- If a required formal representation contains an unresolved structural, dimensional, definitional, or mapping inconsistency, choose "spurious" even when the final scalar answer is correct.
- If it is unclear whether a statement is actually wrong or merely imprecise, and it is not part of a required condition, calculation, definition, formula, unit/scale relation, dimensional relation, mapping, formal representation, check, or final summary, choose "valid" if the reasoning is otherwise sound.
\end{promptbox}

\end{document}